\documentclass[runningheads]{llncs}
\usepackage[T1]{fontenc}

\usepackage{color, colortbl}
\definecolor{Gray}{gray}{0.9}
\definecolor{LightCyan}{rgb}{0.88,1,1}
\usepackage{graphicx}
\usepackage{hyperref}
\usepackage[dvipsnames]{xcolor}
\usepackage{amssymb}
\usepackage{cite}
\usepackage{arydshln}
\usepackage{multirow}
\begin{document}
\title{ A Novel Multi-Stage Prompting Approach for Language Agnostic MCQ Generation using GPT}
\titlerunning{A Novel MSP Approach for Language Agnostic MCQ Generation using GPT}
%
%
\author{Subhankar Maity\orcidID{0009-0001-1358-9534} \and
Aniket Deroy\orcidID{0000-0001-7190-5040} \and
Sudeshna Sarkar\orcidID{0000-0003-3439-4282}}
\authorrunning{S. Maity et al.}
%
\institute{IIT Kharagpur, Kharagpur, India\\ 
\email{\{subhankar.ai,roydanik18\}@kgpian.iitkgp.ac.in},
\email{sudeshna@cse.iitkgp.ac.in}}
%
\maketitle              
\begin{abstract}

We introduce a multi-stage prompting approach (MSP) for the generation of multiple choice questions (MCQs), harnessing the capabilities of GPT models such as  \texttt{text-davinci-003} and GPT-4, renowned for their excellence across various NLP tasks. Our approach incorporates the innovative concept of \textit{chain-of-thought} prompting, a progressive technique in which the GPT model is provided with a series of interconnected cues to guide the MCQ generation process. Automated evaluations consistently demonstrate the superiority of our proposed MSP method over the traditional single-stage prompting (SSP) baseline, resulting in the production of high-quality distractors. Furthermore, the one-shot MSP technique enhances automatic evaluation results, contributing to improved distractor generation in multiple languages, including English, German, Bengali, and Hindi. In human evaluations, questions generated using our approach exhibit superior levels of \textit{grammaticality}, \textit{answerability}, and \textit{difficulty}, highlighting its efficacy in various languages. 

\keywords{Question Generation (QG) \and Multiple Choice Questions (MCQs) \and Chain-of-Thought(CoT) \and Prompt \and GPT }
\end{abstract}

\section{Introduction and Background}

Multiple choice questions (MCQs) are a common way to assess learner knowledge, but manually creating them is time-consuming and demanding for educators \cite{r49}. This underscores the importance of automating MCQ generation. The prompting approach involves enhancing the performance of a large language model (LLM) in downstream tasks by providing additional information, known as a “prompt” to condition its generation \cite{r48}. Recently, the use of prompts has gained prominence in several natural language generation (NLG) tasks, such as interview question generation (QG) \cite{r8}, summarization \cite{r45}, machine translation (MT) \cite{r47}, etc. These approaches take advantage of state-of-the-art GPT models. However, there is limited exploration of prompt-based GPT models for generating MCQs, which presents a unique challenge in harnessing the full potential of these GPT models. Our work seeks to address this gap by investigating the application of prompt-based GPT models in the creation of MCQs.
Very few works explore prompt-based LLMs for the generation of MCQs in multilingual settings (especially for low-resource languages) \cite{r33}. This approach is crucial to addressing language barriers, promoting accessibility, and advancing education in underserved communities. Inspired by recent advances in \textit{chain-of-thought} (CoT) prompting, we seek to bridge the gap between human-like reasoning and the ability of GPT models to generate MCQs. In the work by Wei et al. \cite{r7}, few-shot CoT prompting demonstrates the potential to explicitly generate intermediate reasoning steps, facilitating a more accurate prediction of the final answer with step-by-step manual reasoning demonstrations. Additionally, zero-shot CoT, as introduced by Kojima et al. \cite{r32}, remarkably improves the performance of LLMs without requiring manually crafted examples by simply appending “\textit{Let's think step by step}” to the prompt. Moreover, a work by Tan et al. \cite{r47} utilizes the multi-stage prompting (MSP) approach to improve the task of MT using GPT. In this paper, we extend the idea of CoT to the domain of MCQ generation using MSP \cite{r35}. Our approach involves four essential stages: paraphrase generation, keyword extraction, QG, and distractor generation. By guiding the GPT models through these sequential interactions, we encourage iterative reasoning, which creates well-crafted MCQs. By extending MSP to multiple languages, our approach enables GPT models to create linguistically diverse MCQs. 

Our contributions are as follows: (i) We propose a lightweight language agnostic multi-stage prompting (MSP) approach to generate zero-shot MCQs using GPT models such as \texttt{text-davinci-003} \cite{r36} and GPT-4 \cite{r37} without fine-tuning the models. We also try the one-shot technique in our proposed method to improve the accuracy of our method; (ii) Since automated metrics have their own limitations, in terms of evaluating questions \cite{r44}, we perform a human evaluation of the generated questions.\\ \\
\textbf{State-of-the-Art.} Kumar et al. \cite{r41} proposed a hybrid method using ontology-based and machine learning-based techniques to automatically generate MCQ stems for educational purposes.  Hadifar et al. \cite{r40} fine-tuned T5  \cite{r14} for the generation of educational MCQs in their proposed EduQG dataset. Rodriguez-Torrealba et al. \cite{r9} used fine-tuned T5 for the generation of questions, answers, and associated distractors. Wang et al. \cite{r42} explored the use of a Text2Text formulation to generate cloze-style MCQs using LLMs such as BART \cite{r12} and T5. Vachev et al. \cite{r43} presented a system called Leaf, which used T5 to generate questions, answers, and distractors from the given text. Kalpakchi et al. \cite{r22} explored the ability of GPT-3 to create Swedish prompt-based MCQs in a zero-shot manner. All these prior works (except \cite{r22}) focused on the generation of English MCQs. Moreover, no existing MCQ generation research explores the capability of GPT models, such as \texttt{text-davinci-003} and GPT-4, for prompt-based MCQ generation in multiple languages.

\section{Method}

\begin{figure}[h!]
  \centering
  \includegraphics[width=1\linewidth]{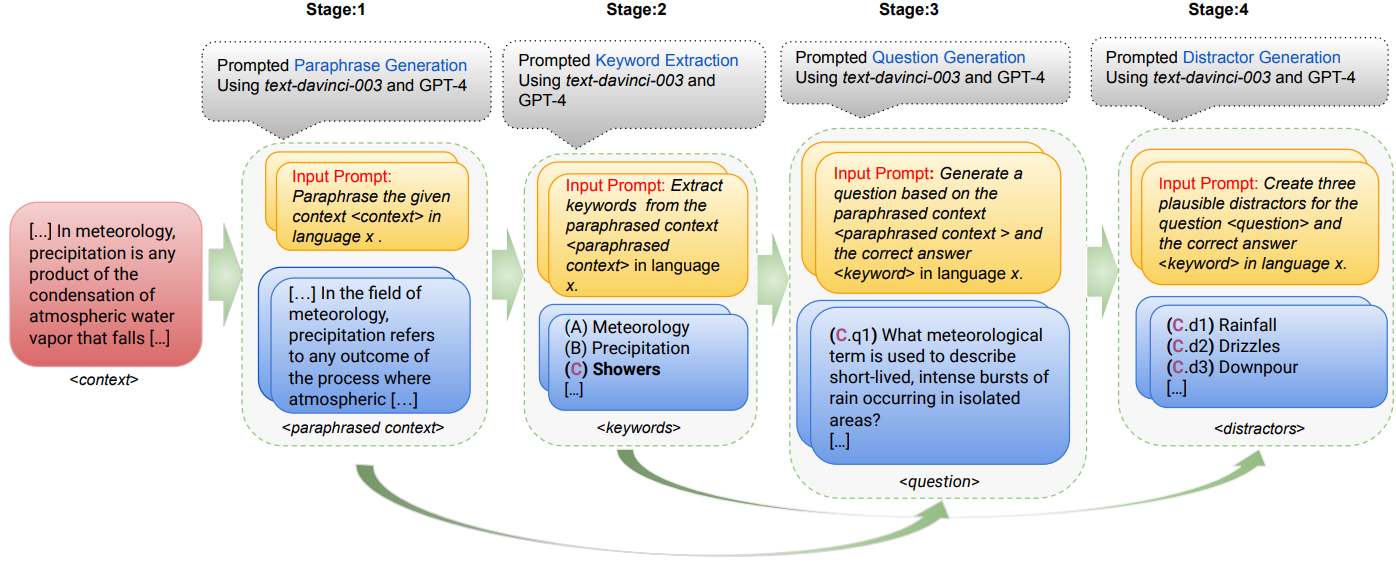}
  \caption{Overview of proposed MSP approach for MCQ generation in various languages. $ x \in \{\textbf{English}, German, Hindi, Bengali\}$. In this example, $x = \textbf{English}.$} \label{fig1}
\end{figure}

Given a set of contexts C = \{$c_1$, $c_2$, ..., $c_N$\}, the goal is to generate diverse and high-quality MCQs for each context $ c_i \in C $ along with their corresponding answer options.
We aim to demonstrate the efficacy of our proposed prompt-based MCQ generation approach using \texttt{text-davinci-003} (abbreviated as \texttt{Davinci}) and GPT-4 to provide high-quality, diverse, and language-agnostic MCQs. \\ \\
\textbf{(A) Multi-Stage Prompting (MSP) Approach}\\ 
The proposed method called MSP (see Fig. \ref{fig1}) involves four steps as follows:\\ \\
\textbf{Paraphrase Generation(PG).}
We aim to generate multiple paraphrases for each context $c_i \in  C$ to ensure diversity in the QG process. Let $P_i = \{p_{i1}, p_{i2}, ..., p_{iM}\}$ be the set of paraphrases for context $c_i$, where M is the number of paraphrases generated. The prompt we use M (M = 3) times is “\textit{Paraphrase the given context <context> in language x}”, where $ x \in \{English, German, Hindi, Bengali \}.$\\ \\
\textbf{Keyword Extraction(KE).}
Next, we extract keywords from each paraphrase $p_{ij} \in P_i$ to represent important information in the generated questions. Let $K_{ij} = \{k_{ijk} | 1 \leq k_{ijk}\leq V\} $ be the set of extracted keywords for paraphrase $p_{ij}$, where V represents the size of the vocabulary. The prompt we utilize is, “\textit{Extract keywords from the paraphrased context <paraphrased context> in language x}”.\\ \\
\textbf{Question Generation(QG).}
Using the extracted keywords $K_{ij}$ for each $p_{ij}$, our objective is to create questions that capture the essential information from the paraphrased context $p_{ij}$. Let $Q_{ij} = \{q_{ijk} | 1 \leq q_{ijk} \leq Q\} $ be the set of questions generated for paraphrase $p_{ij}$, where Q is the number of questions generated for each paraphrase. We feed the prompt as “\textit{Generate a question based on the paraphrased context <paraphrased context> and the correct answer <keyword> in language x}”. In addition, we have tried the one-shot technique to improve the quality of the question. The one-shot technique for QG involves the following prompt: “\textit{For the paraphrased context <paraphrased $context_i$> and the correct answer <$keyword_i$>, the question is <$Question_i$> in language x. Generate a question based on the paraphrased context <paraphrased $context_j$> and the correct answer <$keyword_j$> in language x}”.\\ \\
\textbf{Distractor Generation (DG).}
To form plausible answer options for the generated questions, we introduce the generation of distractors using relevant information from the corresponding question and keyword. Let $D_{ijk} = \{d_{ijkl} | 1 \leq d_{ijkl} \leq D\}$ be the set of distractors for question $q_{ijk}$, where D (D = 3) represents the number of distractors per question. The prompt we use is, “\textit{Create three plausible distractors for the question <question> and the correct answer <keyword> in language x}”. 
In addition, we have tried the one-shot technique to improve the quality of the distractors. The one-shot technique for DG involves the following prompt: “\textit{The distractors for the question <$question_i$> and the correct answer <correct $answer_i$> are <$distractor_{i1}$, $distractor_{i2}$, $distractor_{i3}$> in language x. Create three plausible distractors for the question <$question_j$> and the correct answer <$keyword_j$> in language x}”. \\ 


\noindent \textbf{(B) Baseline: Single-Stage Prompting (SSP) Approach} \\
 There is existing work \cite{r9} for the English language in which a T5 model is fine-tuned on the DG-RACE dataset \cite{r50} to generate distractors for MCQs. However, there is no existing work in multilingual settings, such as German, Hindi, and Bengali, where an encoder-decoder-based model is explored for DG. 
 
 Therefore, we compare our proposed MSP approach, inspired by CoT prompting \cite{r7} and involving multiple reasoning steps, with the basic (single-stage) prompting approach to investigate the effectiveness of our method. Here, the prompt we use is, “\textit{Generate MCQs based on the given context <context> along with the correct answer and three distractors in language x}”.

\section{Experiments}
We evaluate our proposed MSP-based approach in zero-shot and one-shot settings. We compare the results of proposed MSP approach with those of SSP baseline in a zero-shot setting. Additionally, we rigorously examine the effectiveness of our approach in multilingual scenarios, incorporating both high-resource languages such as English and German, as well as low-resource languages such as Hindi and Bengali. This section covers details about the dataset used in our experiments followed by details of the implementation with evaluation metrics. \\ \\
\textbf{Dataset.} Generally, the datasets used for the QG task are taken from the datasets used for question answering (QA). These QA datasets include triples of <\textit{Context}, \textit{Question}, \textit{Answer}>. We use \textit{SQuAD} \cite{r38} as the English (En) MCQ generation dataset. It consists of more than 100K carefully crafted questions in English. We use \textit{GermanQuAD} \cite{r4} as the German (De) MCQ generation data. It consists of more than 13K human-generated QA pairs in German Wikipedia articles. \textit{HiQuAD} \cite{r5} is a special dataset to generate questions in Hindi (Hi). It consists of 6,555 pairs of questions and answers. The questions were created based on a Hindi storybook. We use \textit{BanglaRQA} \cite{r6} as Bengali (Bn) MCQ generation data. It consists of more than 14K human-generated QA pairs in Bengali Wikipedia articles. We randomly sampled 850 contexts from each of the above four QA datasets for our experiments. \\ \\
\textbf{Experimental Setup.}
 We use \texttt{Davinci} with \textit{max tokens} = 50, \textit{presence penalty} = 1.0, \textit{frequency penalty} = 0.0, and \textit{temperature} = 0.7. We use GPT-4 with \textit{maximum tokens} = 50 and \textit{temperature} = 0.7. We use the paid APIs of the GPT models\footnote{\url{https://platform.openai.com/docs/models}}. We use the 850 samples present in a particular language for automatic evaluation. Following the approach of Rodriguez-Torrealba et al. \cite{r9}, we evaluate the quality of the generated distractors using several automated evaluation metrics, including BLEU (from 1 to 4 grams) \cite{r46}, ROUGE-L \cite{r25} and cosine similarity (CS) based on mBERT embeddings \cite{r34}. For each sample, we calculate scores by comparing the correct answer with three generated distractors. This process is then repeated for all correct answers generated by the MSP (Proposed) and SSP (Baseline) approaches from each context in our samples. Finally, we report the average scores for all pairs of <correct answer, distractor> in a particular language in Table \ref{tab:t1}. Our implementation code and datasets will be available at \url{https://github.com/my625/CoT-MCQGen}. \\
\begin{table}
\caption{Automatic evaluation results for various GPT models for DG. We selected the GPT model that performs best among \texttt{\textcolor{purple}{Davinci}} and \texttt{\textcolor{teal}{GPT-4}} in 0-shot (w/ MSP) and then explored the same model in 0-shot (w/ SSP) and 1-shot (w/ MSP). Each cell in this table represents the average value of BLEU-1, BLEU-2, BLEU-3, BLEU-4, ROUGE-L, and cosine similarity (CS) based on mBERT embeddings between every <correct answer, distractor> pair generated by the corresponding GPT model and \textcolor{blue}{language} setting. A higher average value denotes the better quality of the generated distractors. Best results are marked in \textbf{bold} and values marked with * are statistically significant on performing student \textit{t}-test at the 95\% confidence interval w.r.t the best performing GPT model (w/ SSP). }
\label{tab:t1}    
    \centering
    \renewcommand{\arraystretch}{1}
\scalebox{0.65}{
\begin{tabular}{p{0.35\textwidth}|l|l|l|l|l|l|l}
\hline
 \textbf{Model (Approach)} & \textbf{Setting} & \textbf{BLEU-1} & \textbf{BLEU-2} & \textbf{BLEU-3} & \textbf{BLEU-4} & \textbf{ROUGE-L} & \textbf{CS} \\ \hline



\multicolumn{8}{c}{\textcolor{blue}{English}}          \\ \hline

\texttt{\textcolor{purple}{Davinci}} (w/ MSP)&0-shot& 14.75 &7.68 &\textbf{4.47} & 2.47& 13.47& 0.73*\\

\texttt{\textcolor{teal}{GPT-4}} (w/ MSP) & 0-shot & 14.47  & 7.45 & 4.28 & 2.27 & 13.18 & 0.69  \\

 \hdashline
\texttt{\textcolor{purple}{Davinci}} (w/ SSP) (Baseline)& 0-shot & \textbf{14.92} & 7.17 &4.45 &2.29& 12.94 & 0.68 \\ \hdashline

\texttt{\textcolor{purple}{Davinci}} (w/ MSP)& 1-shot & 14.64& \textbf{7.69} & 4.15& \textbf{2.49} & \textbf{13.54} & \textbf{0.74*}\\ \hline 

\multicolumn{8}{c}{\textcolor{blue}{German}}     \\ \hline

\texttt{\textcolor{purple}{Davinci}} (w/ MSP) & 0-shot& 12.14* &  \textbf{6.79}&3.77 & 2.58 & 12.88& 0.70 \\
\texttt{\textcolor{teal}{GPT-4}} (w/ MSP)&0-shot  &11.45 & 6.45 & 3.48 & 2.11 & 12.12 & 0.66 \\
 \hdashline
\texttt{\textcolor{purple}{Davinci}} (w/ SSP) 
 (Baseline)&0-shot & 10.89&6.11 &\textbf{3.79} & 2.43 & 12.49& 0.69 \\ \hdashline
\texttt{\textcolor{purple}{Davinci}} (w/ MSP)&1-shot & \textbf{12.15*} & 6.42 & 3.78&\textbf{2.79} & \textbf{12.98}  & \textbf{0.71}\\ \hline

\multicolumn{8}{c}{\textcolor{blue}{Hindi}}         \\ \hline
\texttt{\textcolor{purple}{Davinci}} (w/ MSP)& 0-shot & 9.58* & 4.98 & 2.13 & 1.36& 11.12 & 0.64*\\
\texttt{\textcolor{teal}{GPT-4}} (w/ MSP)&0-shot &9.74  & 5.01&2.44 & 1.59& 11.82& 0.66*  \\ \hdashline
\texttt{\textcolor{teal}{GPT-4}} (w/ SSP) (Baseline)& 0-shot &8.14 & 4.27 & 2.48 & 1.08& 11.28 & 0.60\\ \hdashline
\texttt{\textcolor{teal}{GPT-4}} (w/ MSP)& 1-shot & \textbf{10.47*} & \textbf{5.22*} &  \textbf{3.47*} & \textbf{1.81} & \textbf{12.47} & \textbf{0.67*}  \\ \hline

\multicolumn{8}{c}{\textcolor{blue}{Bengali}}           \\ \hline
\texttt{\textcolor{purple}{Davinci}} (w/ MSP)& 0-shot & 9.44 & 5.41 & \textbf{3.97*} & 2.41& 9.45 & 0.63*\\
\texttt{\textcolor{teal}{GPT-4}} (w/ MSP) & 0-shot & 9.21 & 4.98 & 2.88& 2.31& 9.19&0.62* \\ \hdashline
\texttt{\textcolor{purple}{Davinci}} (w/ SSP) (Baseline) & 0-shot & 8.45 & 4.81 & 2.94 & 2.19 & 9.34 & 0.59 \\ \hdashline
\texttt{\textcolor{purple}{Davinci}} (w/ MSP)&1-shot & \textbf{9.70*} & \textbf{5.48} & 3.87* &\textbf{2.97} &\textbf{11.78*} & \textbf{0.67*}\\ \hline

    \end{tabular}
}

\end{table}

\begin{figure}[h!]
  \centering
  \includegraphics[width=1\linewidth]{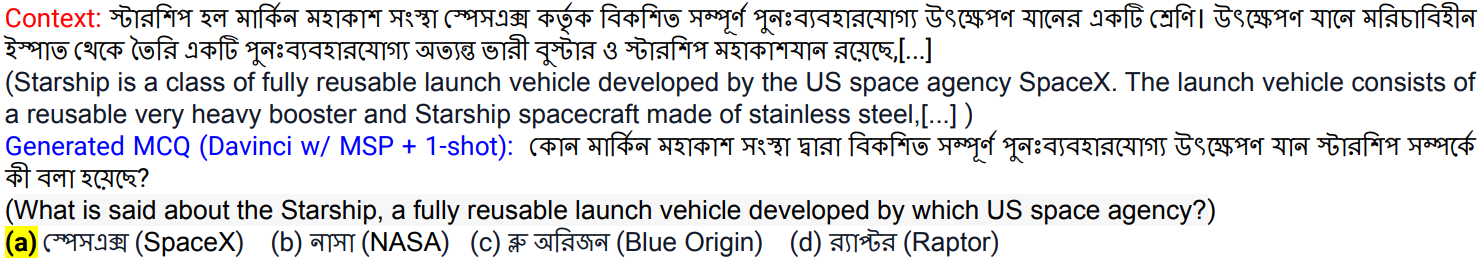}
  \caption{An example of a generated \textit{grammatically} incorrect, low-\textit{answerability} question in Bengali, along with the generated highlighted correct answer \colorbox{yellow}{option} and associated distractors.}\label{fig2}
\end{figure}

\begin{table}[h!]
\caption{Human evaluation results of various best-performing GPT models (w/ MSP) for QG across four languages.}
    \label{tab:t3}
    \centering
    \renewcommand{\arraystretch}{1}
\scalebox{0.65}{
\begin{tabular}{l|l|l|l|l}
\hline
\textbf{Language (GPT Model)} & \textbf{Setting} & \textbf{Grammaticality} & \textbf{Answerability} & \textbf{Difficulty} \\ \hline

\multirow{2}{*} {English (\texttt{\textcolor{purple}{Davinci}})} & 0-shot & 4.38  & 4.29 & 3.68  \\ 

 & 1-shot & \textbf{4.47}  & \textbf{4.42} & 3.75  \\ \hline

\multirow{2}{*} {German  (\texttt{\textcolor{purple}{Davinci}})} & 0-shot & 4.21 & 4.30 & 3.78 \\

 & 1-shot & 4.32 & 4.37 & \textbf{3.89}\\ \hline

\multirow{2}{*} {Hindi (\texttt{\textcolor{teal}{GPT-4}})} & 0-shot & 3.21 & 3.04 & 2.62 \\

 & 1-shot & 3.34 & 3.17 & 2.78 \\ \hline

\multirow{2}{*} {Bengali (\texttt{\textcolor{purple}{Davinci}})} & 0-shot & 3.10 & 2.78 & 2.54 \\

 & 1-shot & 3.27 & 2.98 & 2.62 \\ \hline


    \end{tabular}

}
\end{table}
\noindent \textbf{Results.}
Table \ref{tab:t1} demonstrates the impressive performance of \texttt{Davinci} (w/ MSP) in almost all automated evaluation metrics for one-shot DG tasks in English, German, and Bengali. However, for the Hindi language, GPT-4 (w/ MSP) surpasses all others, exhibiting the best performance across all automated metrics in the one-shot setting. Among all languages, English (w/ \texttt{Davinci}) emerges as the best performer in all automated metrics. Furthermore, German (w/ \texttt{Davinci}) shows a commendable performance. However, Hindi (w/ GPT-4) and Bengali (w/ \texttt{Davinci}) have exhibited lower results in automated metrics compared to other languages. This can be attributed to the fact that most of the \texttt{ Davinci} and GPT-4's training data come from high-resource languages such as English, German, etc. Interestingly, for Hindi and Bengali, the improvement in the one-shot setting (w/ MSP) is higher than that observed in English and German in most cases. It shows that our proposed MSP approach, whether using \texttt{Davinci} or GPT-4, generates higher quality distractors than the SSP approach for the four languages. \\ \\
\textbf{Human Evaluation.}
Taking into account the limitations associated with automated metrics in terms of the evaluation of generated questions \cite{r44}, we conduct a human evaluation by appointing an educated native speaker for each respective language. Each human evaluator\footnote{We use \href{https://www.surgehq.ai/faq}{\textcolor{blue}{Surge AI}} as our evaluation platform.} was asked to rate a total of 400 questions, considering the best-performing GPT model (both for the zero-shot and one-shot settings). The rating scale used ranged from 1 (worst) to 5 (best) based on three criteria: \textit{grammaticality} \cite{r28}, \textit{answerability} \cite{r28}, and \textit{difficulty} \cite{r22}. As shown in Table \ref{tab:t3}, English (w/ \texttt{Davinci}) showed the best performance for \textit{grammaticality} and \textit{answerability}. And for \textit{difficulty}, German ( w/ \texttt{Davinci}) outperforms other languages. On the other hand, Hindi (w/ GPT-4) and Bengali (w/ \texttt{Davinci}) showed poorer performance than high-resource languages such as English and German. This is because GPT models\footnote{\href{https://github.com/openai/gpt-3/blob/master/dataset_statistics/languages_by_word_count.csv}{Hyperlink to a CSV file containing training data statistics for GPT-3} } such as \texttt{text-davinci-003} and GPT-4 have been pre-trained on smaller amounts of data taken from low-resource languages, such as Hindi and Bengali. We also observed improvements in \textit{grammaticality}, \textit{answerability}, and \textit{difficulty} across all languages in the one-shot setting compared to the zero-shot setting. However, the improvement in \textit{grammaticality}, \textit{answerability}, and \textit{difficulty} is greater for low-resource languages, such as Hindi and Bengali, than for high-resource languages, such as English and German. This occurs because showing a one-shot example to the GPT model helps to improve human evaluation criteria more for low-resource languages than for high-resource languages. This is due to the fact that GPT models have been pre-trained on a smaller amount of data from low-resource languages, as discussed earlier. Fig. \ref{fig2} shows an example of a \textit{grammatically} incorrect question with lower \textit{answerability}, along with its options generated by \texttt{Davinci} (w/ MSP) in one-shot setting in Bengali. 

\section{Conclusion and Future Work}

GPT-4 and \texttt{text-davinci-003}, when used with the MSP method, show promise in generating MCQs in multiple languages. They demonstrate potential benefits in terms of performance and quality, both for high-resource and low-resource languages. Our proposed MSP approach outperforms the baseline SSP method, indicating that the MSP approach effectively produces higher-quality distractors. Additionally, the one-shot technique enhances the automatic evaluation results for low-resource languages, addressing the challenges in MCQ generation across various languages. In human evaluation, we observed that the questions generated for high-resource languages exhibit better \textit{grammaticality}, \textit{answerability}, and \textit{difficulty}. Furthermore, the one-shot setting improves the \textit{grammaticality}, \textit{answerability}, and \textit{difficulty} of the generated MCQs compared to the zero-shot setting in all languages. However, further research and fine-tuning may be necessary to improve results for low-resource languages and narrow the gap with high-resource languages in terms of automated and human evaluation metrics.

%
%

\bibliographystyle{splncs04}
\bibliography{ref.bib}

\end{document}